# Ovarian Cancer Detection based on Dimensionality Reduction Techniques and Genetic Algorithm

Ahmed Farag Seddik[1]
[1]Head of Bio-Medical Dep.
Faculty of Engineering, Helwan University
Cairo, Egypt
E-mail address: Ahmed_Sadik@h-eng.helwan.edu.eg
Mobile phone: + 20100-1426485

Hassan Mostafa Ahmed[2]
[2]Master student at Bio-Medical Dep.
Faculty of Engineering Helwan University
Cairo, Egypt
E-mail address: Hmah_hassan@yahoo.com
Mobile phone: +201019156152, +201285472257

**Abstract**

Ovarian cancer presents a late clinical stage in more than 80% of the patients and is associated by a 5-year survival of 35% in this population. Also it is considered to be the 5th most frequent cancer amongst women.
By contrast, the 5-year survival for patients with stage I ovarian cancer exceeds 90%, and most patients are cured of their disease by surgery alone without the need of chemotherapy. Therefore, by increasing the number of detected women in stage I and diagnosing them should have a direct impact on their survival by surgical treatment without the chemotherapy treatment.
Hence, new technologies and features selection techniques for the detection of the early stage ovarian cancer are urgently need to give rise to percentage of the survival.
Pathological changes within organ might be reflected in proteomic patterns in the blood serum. Therefore these changes appearing in the patterns can be detected to differentiate between the cancerous and normal patients.
Here, we aim to select the most significant features that counts for the detection of the ovarian cancer by many features selection techniques, and then evaluating these techniques by calculating the accuracy of each by the aid of classification methods.

In this research, we have two serum SELDI (surface-enhanced laser desorption and ionization) mass spectra (MS) datasets to be used to select features amongst them to identify proteomic cancerous serums from normal serums. Features selection techniques have been applied and classification techniques have been applied as well. Amongst the features selection techniques we have chosen to evaluate the performance of PCA (Principal Component Analysis ) and GA (Genetic algorithm), and amongst the classification techniques we have chosen the LDA (Linear Discriminant Analysis) and Neural networks so as to evaluate the ability of the selected features in identifying the cancerous patterns.

Results were obtained for two combinations of features selection techniques and classification techniques, the first one was PCA+(t-test) technique for features selection and LDA for accuracy tracking yielded an accuracy of 93.0233 % , the other one was genetic algorithm and neural network yielded an accuracy of 100%.

So, we conclude that GA is more efficient for features selection and hence for cancerous patterns detection than PCA technique.

## I. INTRODUCTION

Application of new technologies for detection of ovarian cancer could have an important effect on public health, but to achieve this goal, specific and sensitive molecular markers are essential. This need is especially urgent in women who have a high risk of ovarian cancer due to family or personal history of cancer, and for women with
A genetic pre-disposition to cancer due to abnormalities in pre-disposition genes such as BRCA1 and BRCA2.

Ovarian cancer presents at a late clinical stage in more than 80% of patients,1 and is associated with a 5-yearsurvival of 35% in this population.
By contrast, the5-year survival for patients with stage I ovarian cancer exceeds 90%, and most patients are cured of their disease by surgery alone. Therefore, by increasing the number of detected women in stage I and diagnosing them should have a direct impact on their survival by surgical treatment without the chemotherapy treatment.

Cancer antigen 125 (CA125) is the most widely
used biomarker for ovarian cancer.1–6 Although concentrations of CA125 are abnormal in about 80% of patients with advanced-stage disease, they are increased in only 50–60% of patients with stage I ovarian cancer.CA125 has a positive predictive value of less than 10% as a single marker, but the addition of ultrasound screening to CA125 measurement has improved the positive predictive value to about 20%.

Low-molecular-weight serum protein profiling might reflect the pathological state of organs and aid in the early detection of cancer. Matrix-assisted laser desorption and ionization time-of-flight (MALDI-TOF)and surface-enhanced laser desorption and ionization time-of-flight (SELDI-TOF) mass spectroscopy can profile proteins in this range. These profiles can contain thousands of data points, necessitating sophisticated analytical tools.

We aimed to link SELDI-TOF spectral analysis with a high-order analytical approach using samples from women with a known diagnosis to define
an optimum discriminatory PROTEOMICPATTERN. We then aimed to use this pattern to predict the identity of masked samples from unaffected women, women with early-stage and late-stage ovarian cancer, and women with benign disorders.

Mass spectrometry serum profiles were used to detect early ovarian cancer, also the GA was applied and for identifying ovarian cancer, this technique gave sensitivity of 100% and specifity of 95%.

## II. METHODS AND DATASETS

**Mass Spectrometry**

Mass spectrometry analysis has been discovered by (sirJ.J. Thomson) in the early part of the 20th century. Mass spectrometry (MS) is a technique for 'weighting' individual molecules, fragments of molecules or individual atoms that have been ionized. The basic measurement unit of the MS is the mass-to-change ratio (M/Z). The output pattern consists of an array of M/Z values and corresponding intensity values in another array.
There are two methods for profiling a population of protein in a sample according to the size and net electrical charge of the individual proteins, which are:
   a. MALDI-TOF.
   b. SELDI-TOF.

**MALDI-TOF and SELDI-TOF**

They are methods for profiling a population of proteins in a sample according to the size and net electrical charge of the individual proteins. The readout is a spectrum of peaks. The position of an individual protein in the spectrum corresponds to its "time of flight" because the small proteins fly faster and the large proteins fly more slowly. SELDI( surface-enhanced laser desorption and ionization time-of-flight )—a refinement of MALDI( Matrix-assisted laser desorption and ionization time-of-flight )—preselects the proteins in the sample by allowing them to bind to the treated surface of a metal bar, which is coated with a specific chemical that binds a subset of the proteins within the serum sample.

**PROTEOMIC PATTERN**

The discriminating pattern formed by a small key subset of proteins or peptides buried among the entire repertoire of thousands of proteins represented in the sample spectrum. The pattern is defined by the peak amplitude values only at key mass/charge (M/Z) positions along the spectrum horizontal axis. For the discrimination of ovarian cancer, the pattern is formed by the combination of spectral amplitudes at five precise M/Z values.

**Datasets**

The spectra found in the data sets we are working on are obtained using SELDI-TOF technique where serum samples were thawed, added to a C16hydrophobic interaction protein chip, and analyzed on the Protein Biology System 2 SELDI-TOF mass spectrometer.

The dataset was downloaded from the FDA-NCI Clinical Proteomics Program Databank.

There are two high resolution datasets that are involved in this research, generated by WCX2
protein array and nci-pbsii protein chip, respectively:
   a. **OvarianCD_PostQAQC Data set:**
      Obtained using the WCX2 protein chip, and consisting of 95 normal and 121 diseased patterns, where every pattern consists of 15,156 M/Z values with corresponding intensity values.

   b. **Ovarian-cancer-nci-pbsii-Data set.**
      Obtained using nci-pbsii protein chip, and consisting of 91 normal and 162 cancerous patterns, also every pattern consists of 15,156 M/Z values with corresponding intensity values.

**Dimensionality Reduction and Features selection techniques**

Dimensionality reduction techniques are used when we have a large number of features to select amongst them the most significant ones to be used for further classification of a certain case.
Dimensionality reductions are used for improving the efficiency of a certain classification algorithm, also used for handling high dimensional data where hundreds of thousands of features are present.

Two major types of dimensionality reduction techniques are present, which are the filteration technique and the wrapping technique.

   a. **Filteration technique**

It is used to reduce the features from few thousands to few hundreds of them, as an example of this technique; the PCA technique which allows us to know the significant features by the method of eigen values and eigen vectors, where the features vector with highest score of eigen value is the most significant one.
Another example is the t-test method where every point of mass spectrum curve was observed as a feature and the equivalent ion
intensity as its value. The significant features were selected by the calculation of mean intensity values for every point in mass spectra of cancer and control groups.

T-test considered each feature independently.
It was assumed that both groups of data values were distributed normally and had similar variances. Test statistics were calculated as follows in equation.1.

$$t = \frac{\overline{x_a} - \overline{x_e}}{\sqrt{(\frac{v_a}{n_a} + \frac{v_e}{n_e})}}$$

Where ;
$\overline{x_a}$ and $\overline{x_e}$ are the mean values of the cancerous and normal intensities respectively.
$v_a$ and $v_e$ are the variances of the two distributions.
$n_a$ and $n_e$ are the number of instances in each distribution.

### b. Wrapping technique

It is further used with the filtration technique to track the accuracy of the selected features with respect to their number, allowing us to reduce these features to few tens of features to be used to train any classifier to identify between two classes.

As an example of this technique; the LDA technique where we use the classifier firstly with the one or two features and then incrementing the number of features involved in classification and computing the accuracy each time we increment the features and then we plot the graph of the accuracy versus the number of features and we get the optimum number of features that the accuracy will be degraded after them.

**Genetic algorithm**

It is considered to be an optimization algorithm for large number of features. The advantage of this algorithm is that it gives the exact discriminative features for the case of study. It applies features reduction and wrapping together to obtain the significant features, so it can not be considered either a filtration technique or a wrapping technique.

### III. PROPOSED TECHNIQUES

We have developed two techniques for obtaining the significant features for identifying the ovarian cancer disease.

### A. First proposed technique

The first technique is to reduce the features using the PCA reduction technique after ranking the features by t-test method, after reducing the features we introduce them to a wrapper technique which is the LDA. Having doing so we choose the most optimum features according to the accuracy curve information.
The following figure illustrates more efficiently.

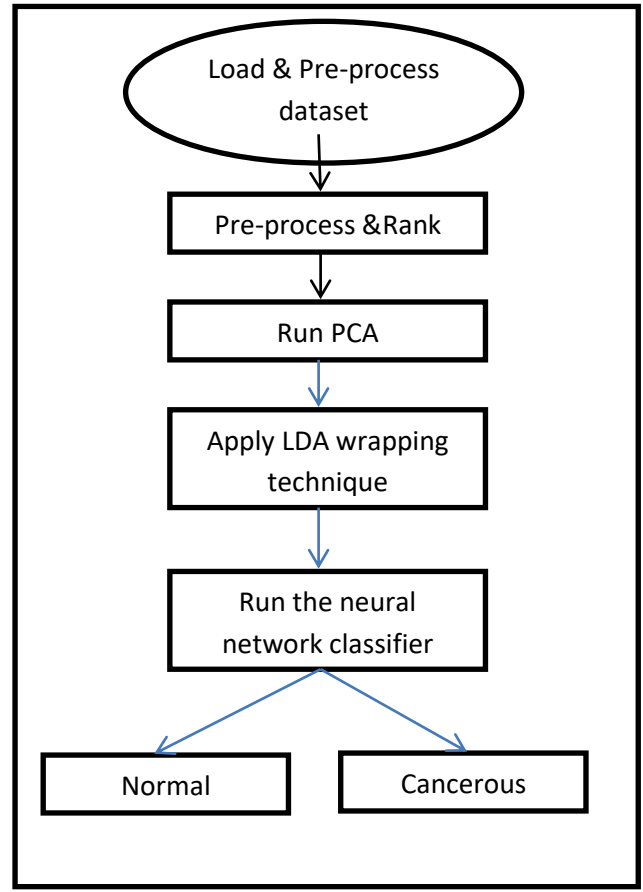

*Figure(1) The first proposed technique.*

### 1. Loading the data and pre-processing step

The data is loaded on Matlab program and then some few samples are observed via plotting tool, as the following figure.

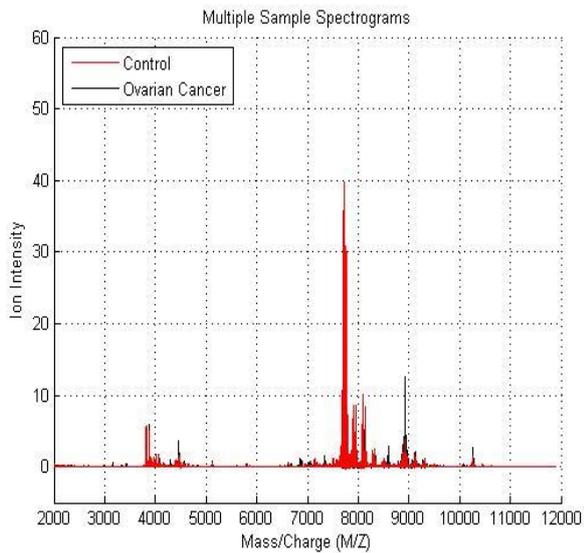

*Figure(2)Mass spectra of cancerous and normal patients.*

We have to mention the following table to make the reader aware of the basic information from each dataset as we got it from the website.

| Dataset | Cancerous | Normal |
|---|---|---|
| OvarianCD_PostQAQC Data set | 121 | 95 |
| ovarian--nci-pbsii- Data set | 162 | 91 |

*Table(1)Distributed groups of the datasets.*

After observing some of the patterns we go ahead to the pre-processing step where the following processing is done:
1. Resampling the spectra.
2. Baseline correction.
3. Normalization.
4. Peak preserving noise reduction.

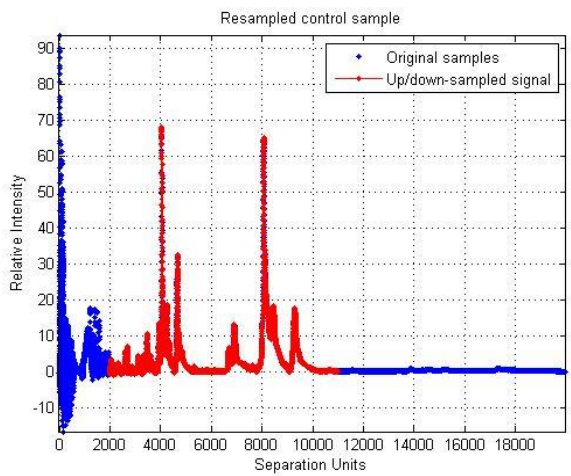

Figure(3) Resampled control sample.

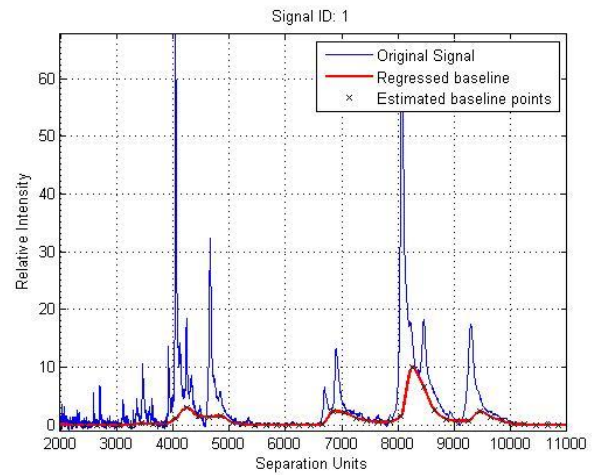

Figure(4) Correct baseline versus the original signal.

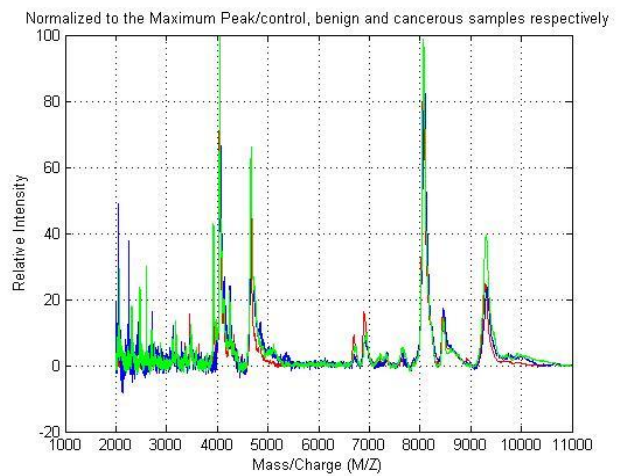

Figure(5) Samples patterns normalized to maximum peak.

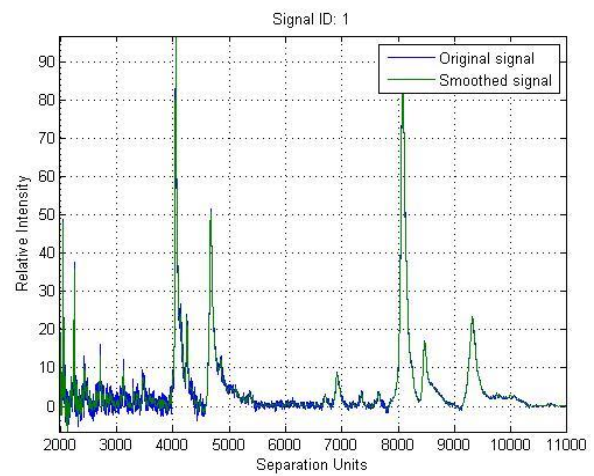

Figure(6) Peak pre-serving.

**2. Features ranking step**

In this step we do a partitioning to the group which contains the cancerous and normal samples by the *Cross Validation* method with the holdout technique. After having the group partitioned to train(173 samples) and test(43 samples) we do a further features ranking for the features found in the training group. Feature ranking is done by many methods each has advantage on features selection that will be done in next step, also this makes difference in calculating accuracy of classification, and this will be observed in the results section. Amongst these methods: t-test, entropy, ROC and Wilcoxon.

### 3. Features selection step

The PCA is applied in this step to reduce features of the training group to few hundreds and then projecting these features in the PCA space to be used in wrapping step.After having the PCA matrix of features we perform wrapping technique using the LDA classifier, where we increment the number of features every time and check for accuracy of classification and plot a graph of accuracy versus the number of features, the following curves illustrates the process of wrapping.

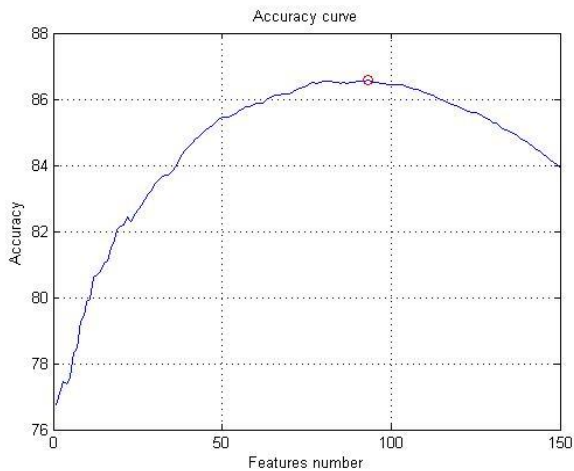

Figure(7)Obtained by t-test ranking method.

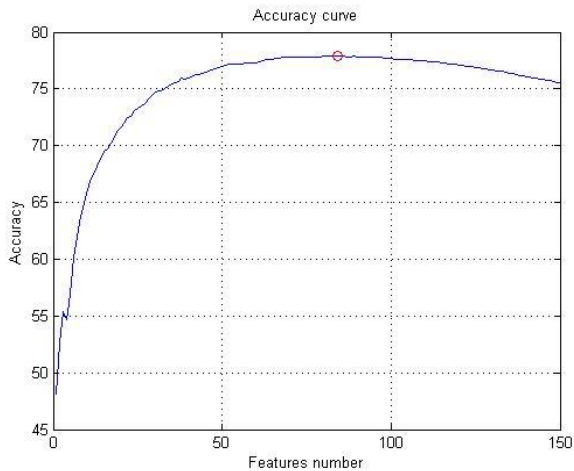

Figure(8)Obtained by Entropy_ranking method.

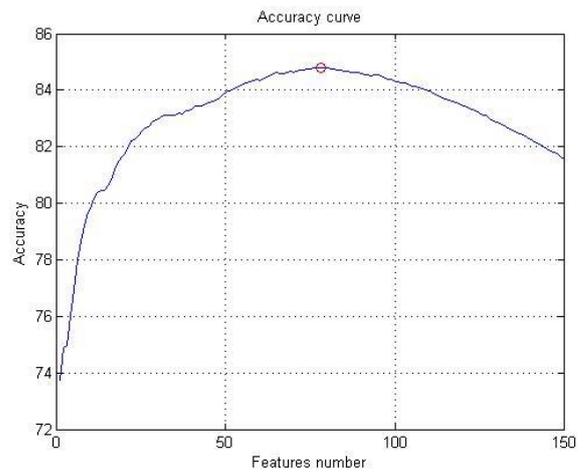

Figure(9)Obtained by ROC ranking method.

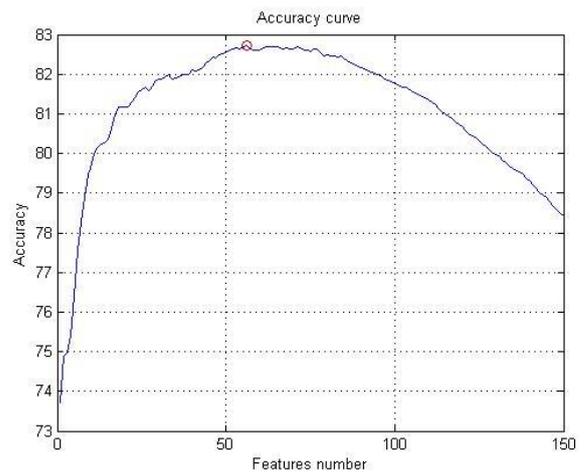

Figure(10)Obtained by Wilcoxon ranking method.

### 4. Results of the 1st proposed technique

In this technique we have obtained many results of percentage correct classification each corresponds to the method of features ranking that is used along with the PCA reduction technique, the results are shown in the following table, and they will be discussed later to show the advantages of each ranking method over the other and its effect on correct classification.

The results found in the table are obtained by applying a Feed forward neural network to classify the cancerous from the normal patients based on the features obtained from the wrapping techniques.

| Ranking Method | Percentage of correct classification |
| --- | --- |
| t-test | 93.0233 % |
| Entropy | 86.0465 % |
| R.O.C | 74.4186 % |
| Wilcoxon | 88.3721 % |

Table(2)Results obtained by different methods of features ranking.

From the results in the table we have to give a brief discussion to make the reader more able to understand how the different methods of features ranking can affect the classification process. First of all, we notice that the highest correct percentage of classification is found with t-test technique, Wilcoxon technique coming in the second position after it and then each other technique according to its result, so we conclude that t-test is most efficient for our study in the 1$^{st}$ proposed technique. However we have to understand that it is a matter of compromise, where we can not say that a specific features ranking technique is the ideal one, otherwise we can say it is the optimum one for the classifier to be used with it.
To understand this; behold the curves well enough and you will see that:
  a. In t-test method; the accuracy has got its maximum values in the middle of the curve; i.e. after a large number of features, while leaving the accuracy degraded in the left region of the curve, so if we are to design a classifier we have to take into account the complexity introduced by increasing the number of features involved in classification.
  b. In Entropy method; the accuracy reached its maximum value and continued in a plateau manner before reaching even the number of features it reached in t-test method, and this gives rise and impact to use a classifier that does not have to face the complexity introduced by the increase of the features number, where the accuracy is almost the same for the plateau region.
  c. In R.O.C method; the curve resembles that of t-test but with low accuracy reached and this accounts for the low percentage of correct classification found in the table, and it is not favoured till you have a low capabilities involved in the classifier.

In Wilcoxon method; we presume it is the optimum method where we observe that the curve combined between the high accuracy found in t-test and the plateau region found in the Entropy method allowing us to choose whether to increase the number of features involved in classification or not without degrading the classification so much, and this gives us the ability to add or discard over fitting features without much loss in the classification.

B. **Second Proposed technique**

In this technique we use the Genetic algorithm as an optimizing technique for reducing the number of features involved in the classification method, after that we evaluate the performance by applying these features to a feed forward neural network.
The following figure illustrates our concept more efficiently.

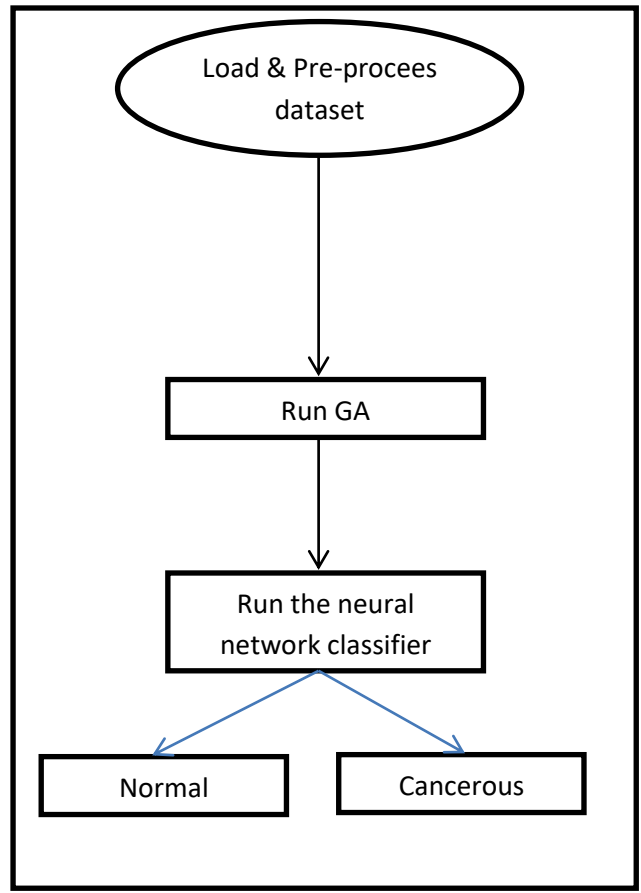

Figure(11)The second proposed technique.

The steps of this technique is as the same as the first one in loading the features and pre-processing them. Having the data pre-processed we turn to apply the genetic algorithm. The genetic algorithm to be applied requires some steps to be done, after them we obtain the identifying features for classification, where we obtain a 100% accuracy, these steps are:

1. **Creating a fitness function for GA**

The GA requires a fitness function that describes the phenomenon that we are about to study and optimize. The fitness function also tests small subsets of M/Z values and determines which to be passed and which to be removed from each subsequent generation. The fitness function biogafit is passed to the genetic algorithm solver using a handle function. Biogafit maximizes the separability of two classes by using a linear combination of:
1. The a-posteriori probability.
2. The empirical error rate of a linear classifier (classify).

2. **Creating an initial population**

Here we can change how the optimization is performed by the genetic algorithm by creating custom functions for crossover, fitness scaling, mutation, selection, and population creation.

### 3. Setting the GA options

The GA function uses an options structure to hold the algorithm parameters that it uses when performing a minimization with a genetic algorithm. The gaoptimset function will create this options structure. For demonstration purposes the genetic algorithm will run only for 50 generations. However, you may set 'Generations' to a larger value.

### 4. Run GA to Find 20 Discriminative Features

Use *ga* to start the genetic algorithm function. 100 groups of 20 data points each will evolve over 50 generations. Selection, crossover, and mutation events generate a new population in every generation.

### 5. Displaying the discriminating features

We then show the discriminating features to observe their locations along the pattern.

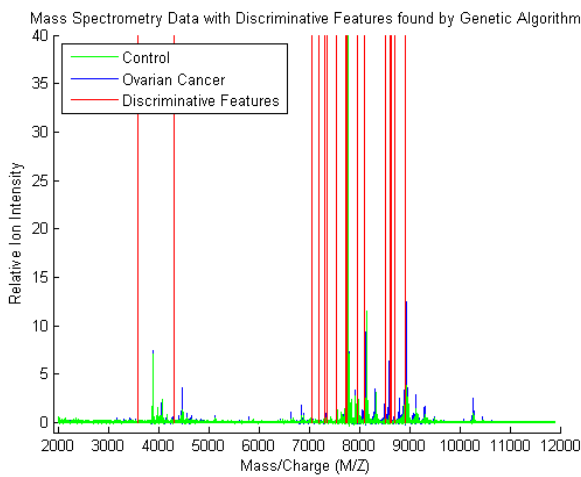

Figure(12)The discriminating features superimposed on the pattern.

### 6. Results of the 2nd technique

The result of this technique is 100% correct classification, and this is possible since we put our hands on the real discriminating features not on the most discriminating features as the other techniques (PCA, LDA) do.

## IV. CONCLUSION

In the 1st technique the accuracy is not 100% because of the lack in the discriminating features because the techniques used to reduce the dimensionality do not allow us to extract all significant features, instead it allows to extract the most discriminating features which means that there might be some features that are not included with us in the classification even if they are over-fitting features and hence the accuracy degradation is observed.

On the other hand the genetic algorithm does offer us the advantage of extracting the exact discriminating features that give rise to a full correct classification, and hence we recommend the genetic algorithm for further features optimization in ovarian cancer field.

REFERENCES

[1] *T.P. Conrads, et al., "High-resolution serum proteomic features for ovarian detection", Endocrine-Related Cancer, 11, 2004, pp. 163-178.*J. Clerk Maxwell, A Treatise on Electricity and Magnetism, 3rd ed., vol. 2. Oxford: Clarendon, 1892, pp.68-73.

[2] *E.F. Petricoin, et al., "Use of proteomic patterns in serum to identify ovarian cancer", Lancet, 359(9306), 2002, pp. 572-577.*R. Nicole, "Title of paper with only first word capitalized," J. Name Stand. Abbrev., in press.

[3] R.H. Lilien, et al., "Probabilistic Disease Classification of Expression-Dependent Proteomic Data from Mass Spectrometry of Human Serum", Journal of Computational Biology, 10(6), 2003, pp. 925-946.

[4] Fabian J. Thies and Anke Mayer-Bäse, ''Biomedical Signal Analysis, Contemporary Methods and Applications'', 9, 2009.

[5] FDA-NCI Clinical Proteomics Program Databankwebsite.http://home.ccr.cancer.gov/ncifdaproteomics/ppatterns.asp.